\documentclass{article} 
\usepackage{iclr2027_conference,times}

\usepackage{amsmath,amsfonts,bm}

\def\eqref#1{equation~\ref{#1}}

\def\1{\bm{1}}

\DeclareMathAlphabet{\mathsfit}{\encodingdefault}{\sfdefault}{m}{sl}
\SetMathAlphabet{\mathsfit}{bold}{\encodingdefault}{\sfdefault}{bx}{n}

\usepackage{amsmath}
\usepackage{amssymb}
\usepackage{amsthm}
\usepackage{mathtools}
\usepackage{algorithm}
\usepackage{algpseudocode}

\usepackage{hyperref}
\usepackage{url}
\usepackage{xspace}
\usepackage{booktabs}
\usepackage{tabularx}
\usepackage{multirow}
\usepackage{graphicx}
\usepackage{fvextra}
\usepackage{textcomp}

\DefineVerbatimEnvironment{ReframePrompt}{Verbatim}{
    fontsize=\small,
    breaklines=true,
    breakanywhere=false,
    breaksymbolleft={},
    commandchars=\\\{\}
}

\newcommand{\evalpair}[2]{%
    \makebox[2.2em][r]{#1}\,/\,\makebox[2.2em][l]{#2}}
\newtheorem{theorem}{Theorem}[section]
\newtheorem{lemma}[theorem]{Lemma}
\newtheorem{proposition}[theorem]{Proposition}
\newtheorem{corollary}[theorem]{Corollary}

\theoremstyle{definition}

\theoremstyle{remark}

\title{\textsc{Timbre}: Reliability-Aware Harmonic Consensus for
Label-Free Multimodal Test-Time Reinforcement Learning}

\author{
Haotian Lyu$^{1,*}$,
Haoyu Wang$^{2,*}$,
Ru Wang$^{3}$,
Feng Ji,
Ziwen Li$^{2}$,
Jiaxian Guo$^{4}$,
Mingming Gong$^{1,2}$
\\[1ex]
$^{1}$University of Melbourne\\
$^{2}$Mohamed bin Zayed University of Artificial Intelligence\\
$^{3}$The University of Tokyo\\
$^{4}$Google\\[1ex]
$^{*}$Equal contribution.
}

\newcommand{\method}{\textsc{Timbre}\xspace}

\iclrfinalcopy 
\begin{document}

\maketitle

\begin{abstract}
Label-free test-time reinforcement learning adapts vision--language
models using rewards derived from their own predictions. However,
misinterpreted visual evidence and sensitivity to textual formulation
can produce a consistent but incorrect consensus that majority voting
reinforces. Cross-view aggregation can expose such errors, but image
and text transformations may themselves degrade task-relevant evidence
to different extents. We introduce \method, a reliability-aware harmonic
consensus framework for multimodal test-time reinforcement learning.
\method constructs a $2\times2$ grid combining original and perturbed
images with direct-solving and reframing modes, selects pseudo-labels
through weighted harmonic aggregation, and adapts modality-level
weights using batch-averaged, temporally smoothed shifts in pseudo-label
confidence as a heuristic reliability signal. 
We formulate consensus as a weighted $\chi^2$-barycenter
problem whose unique minimizer is the normalized weighted harmonic
aggregate, without requiring correct-answer invariance. An exact
support--dispersion decomposition characterizes when harmonic
aggregation preserves or corrects arithmetic rankings, while a
sensitivity analysis establishes how modality reweighting changes
pairwise ranking margins, informing interpretation of the heuristic
updates. Across six benchmarks and three Qwen3-VL model scales,
\method outperforms TTRV in 16 of 18 benchmark--model settings on
adaptation instances and 17 of 18 settings on held-out instances under
matched estimation and optimization rollout counts. These results
demonstrate improved target-task adaptation and transfer to unseen instances without further updates.
\end{abstract}

\section{Introduction}
\label{sec:introduction}

Vision--language models (VLMs) have rapidly advanced in visual
understanding and instruction following
\citep{liu2023llava,chen2024internvl,bai2025qwen3vl}, and are
increasingly applied to multimodal reasoning over diagrams,
charts, and mathematical figures \citep{lu2024mathvista}.
Reinforcement learning with verifiable rewards (RLVR) has
emerged as an effective post-training paradigm for reasoning
\citep{shao2024deepseekmath,guo2025deepseekr1} and has recently
been extended to VLMs \citep{liu2025visualrft,huang2026visionr1}.
However, verifiable answers may be unavailable for unlabeled
inputs at deployment. Label-free test-time reinforcement
learning instead derives rewards from the model's own
predictions: TTRL \citep{zuo2025ttrl} uses majority-vote
pseudo-labels, while TTRV \citep{ttrv2025} extends test-time
RL to VLMs through frequency-based self-rewards. Such intrinsic
reward optimization sharpens existing answer preferences
\citep{he2026how}, making its effectiveness dependent on
alignment between confidence and correctness.

Yet a model's dominant answer need not be correct
\citep{yan-etal-2026-consensus}. In multimodal reasoning,
systematic errors in visual perception, interpretation, or
reasoning \citep{tong2024eyes,zhang2024mathverse,zhou2024paraphrase}
can produce a consistent but incorrect consensus.
Cross-view aggregation offers a way to challenge
presentation-dependent errors, including those arising from
reliance on superficial cues. Self-Harmony
\citep{selfharmony2025} aggregates answer frequencies from
original and reframed questions using a harmonic mean,
penalizing candidates with weak support in either view.
Its theoretical analysis rests on the assumption that semantically equivalent
reframing leaves the correct answer's probability unchanged,
while incorrect-answer probabilities may vary across views.

This assumption is fragile when extending Self-Harmony to
multimodal reasoning. Visual perturbations may obscure essential
details, while textual reframing must preserve both explicit
constraints and the query's grounding in visual entities and
relations. Even when task semantics remain unchanged, a new
presentation can make relevant evidence easier or harder for
the model to use, so the invariance-based justification may
no longer apply. This limitation motivates a consensus
formulation that explicitly controls each view's influence
without requiring correct-answer invariance, with weights
adapted using reliability signals available without
ground-truth answers.

To this end, we seek a consensus distribution that balances
agreement with individual view distributions according to
their assigned reliability weights. We formulate this as
a weighted Pearson $\chi^2$-barycenter problem and show that
its unique minimizer is the normalized weighted harmonic
aggregate. This provides an exact optimization interpretation
of harmonic consensus without requiring correct-answer
invariance.

We then decompose the harmonic score exactly into weighted
arithmetic support discounted by cross-view dispersion.
This decomposition characterizes when harmonic aggregation
preserves a correct arithmetic ranking or corrects an
arithmetic error. Exact invariance of correct-answer support
emerges as a sufficient condition for preserving every strict
arithmetic win of the correct answer. Finally, we derive an
exact sensitivity formula for the effect of modality-level
reweighting on pairwise ranking margins. This formula
characterizes when a weight adjustment increases the margin,
providing an analytical basis for interpreting our
confidence-shift heuristic.

\begin{figure}[ht]
    \centering
    \includegraphics[width=\textwidth]{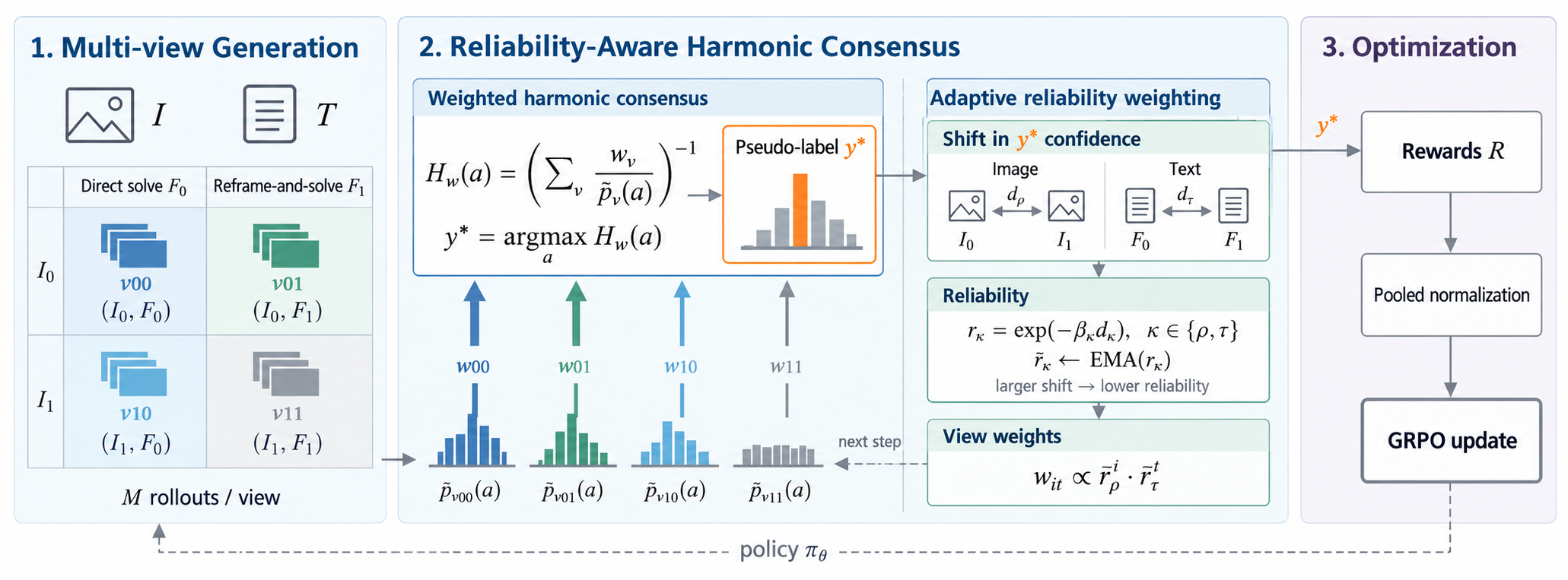}
    \caption{\textbf{Overview of TIMBRE.}
    Four views combine original or perturbed images with
    direct-solving or reframe-and-solve prompts.
    Weighted harmonic consensus selects the pseudo-label $y^*$.
    Cross-view shifts in confidence assigned to $y^*$ update
    modality-level reliability factors, which determine view
    weights for the next step (dashed arrow).
    For each instance, rollout rewards are pooled and jointly
    normalized to compute GRPO advantages.}
    \label{fig:overview}
\end{figure}

Building on this formulation, we introduce \method, a
reliability-aware harmonic consensus framework for label-free
multimodal test-time reinforcement learning
(Figure~\ref{fig:overview}). \method constructs a $2\times2$
grid pairing original and perturbed images with original and
reframed queries, and selects pseudo-labels through weighted
harmonic aggregation of their answer distributions.
Batch-averaged shifts in confidence assigned to the current
pseudo-label across original and transformed views serve as
a heuristic proxy for transformation unreliability.
Larger shifts yield lower target reliability factors, which
are temporally smoothed and combined into modality-factorized
view weights.

Our contributions are threefold:
\begin{itemize}
\item \textbf{Adaptive multimodal consensus.}
We introduce a four-view framework for label-free multimodal
test-time RL that combines weighted harmonic pseudo-label
selection with online modality-level reliability weighting.

\item \textbf{Consensus formulation and exact ranking analysis.}
We formulate consensus as a weighted $\chi^2$-barycenter
problem whose unique minimizer is the normalized weighted
harmonic aggregate. We derive an exact support--dispersion
decomposition, characterize when harmonic aggregation preserves
or corrects arithmetic rankings, and establish how modality
reweighting changes pairwise ranking margins.

\item \textbf{Evaluation under matched rollout budgets.}
Across six multimodal reasoning benchmarks and three model scales in the
Qwen3-VL family \citep{bai2025qwen3vl}, \method outperforms the SOTA
baseline TTRV \citep{ttrv2025} in 16 of 18 benchmark--model settings on
adaptation instances and 17 of 18 settings on held-out instances under
matched estimation and optimization rollout budgets. The held-out results
further demonstrate within-benchmark transfer to unseen instances without
additional updates.
\end{itemize}

\section{Related Work}
\label{sec:related_work}

\paragraph{Label-free test-time reinforcement learning.}
Test-time adaptation updates models on unlabeled test inputs using
objectives such as auxiliary self-supervision \citep{sun2020ttt} and
prediction-entropy minimization \citep{wang2021tent}. Label-free
test-time reinforcement learning instead optimizes the policy with rewards
constructed from its own sampled responses. TTRL \citep{zuo2025ttrl}
rewards agreement with a majority-vote pseudo-label and optimizes the
policy with GRPO \citep{shao2024deepseekmath}. Subsequent variants retain
majority-vote pseudo-labels while improving sampling and optimization:
ETTRL \citep{liu2025ettrl} adapts tree-based rollouts and regularizes
advantages through clipping or entropy-based rescaling, while SPINE
\citep{wu2025spine} combines token-selective updates with entropy-band
regularization to mitigate training collapse. For vision--language
models, TTRV \citep{ttrv2025} uses frequency-based self-rewards with
entropy-based diversity control. These methods differ in how rewards are
shaped and optimized, but derive their learning signals from responses to
a single presentation of each input.

\paragraph{Pseudo-label selection beyond majority voting.}
Self-consistency \citep{wang2023selfconsistency} aggregates answers
across sampled reasoning paths by majority voting, but an incorrect
consensus can reinforce model errors when used as supervision.
SCRL \citep{yan-etal-2026-consensus} improves pseudo-label reliability
through selective positive labeling and entropy-gated negative
supervision, while SCOPE \citep{wang-etal-2026-beyond-majority}
integrates step-wise confidence and subgroup-specific consensus.
Beyond single-input aggregation, Co-Rewarding-I
\citep{zhang2025coreward} derives rewards through cross-supervision
between semantically analogous questions.

Closest to our work, Self-Harmony \citep{selfharmony2025}
aggregates answer frequencies from original and reframed
questions using a harmonic mean. Its analysis formalizes
correct-answer invariance across semantically equivalent views
as $p_{\theta}(C\mid x)=p_{\theta}(C\mid x')$, where $C$ is
the correct answer, and relates harmonic aggregation to a
second-order approximation of a view-invariant Infomax objective
under additional regularity conditions. This invariance
assumption is fragile in multimodal settings, where transformations
can alter access to task-relevant evidence. Because the
Infomax-based justification relies on this assumption, it does
not directly justify reliability-weighted aggregation when
correct-answer probabilities vary across views.

We instead formulate consensus as a weighted Pearson
$\chi^2$-barycenter problem, incorporating view influence
directly into the objective: each weight scales the penalty
for disagreement with the corresponding view's answer
distribution. We show that its unique minimizer is the
normalized weighted harmonic aggregate, providing an exact
optimization basis for our pseudo-label selector without
requiring correct-answer invariance. We adapt the view weights
online using a label-free confidence-shift heuristic.

\section{Method}
\label{sec:method}

Given an unlabeled multimodal test set $\mathcal D$, \method{}
adapts a vision--language model $\pi_\theta$ without ground-truth
answers. Each update generates rollouts from four views, selects
pseudo-labels through weighted harmonic consensus, and optimizes
the policy using jointly normalized cross-view rewards.
Modality-level reliability factors, updated online using a
confidence-shift heuristic, control the influence of transformed
views. Algorithm~\ref{alg:timbre} summarizes the procedure.

\subsection{Multi-View Generation}
\label{sec:method_views}

\paragraph{View construction.}
For each instance $(I,T)\in\mathcal D$, we combine two image
states with two generation modes:
\begin{equation}
    \mathcal V=\{v_{00},v_{01},v_{10},v_{11}\},
    \qquad v_{it}=(I_i,\mathcal F_t(T)),
    \label{eq:view_grid}
\end{equation}
where $I_0=I$, $I_1=g(I)$, and the prompt templates
$\mathcal F_0$ and $\mathcal F_1$ request direct solving and
reframing followed by solving, respectively.
Restricting to $\{v_{00},v_{01}\}$ yields a two-view variant
that applies Self-Harmony's original--reframed construction
to multimodal inputs \citep{selfharmony2025}.
We sample $M$ rollouts per view for distribution and reliability
estimation, while using only the first $K<M$ for policy
optimization. This allocation uses additional forward-only
samples to reduce sampling variability in pseudo-label selection
and reliability estimation, without incurring the additional
backpropagation cost of optimizing on every rollout.

\paragraph{Image transformation.}
The transformation $g$ applies stochastic brightness, contrast,
and saturation adjustments followed by additive Gaussian noise.
Parameters are resampled whenever an instance is drawn, with
images transformed independently for multi-image inputs.
These operations vary low-level appearance while preserving
image coordinates; cropping and large-angle rotation are excluded
to limit disruption of spatial relations and diagram content.
Appearance changes can nevertheless obscure task-relevant details,
motivating adaptive reliability weighting.

\paragraph{Text transformation.}
Following Self-Harmony's fused reframing design
\citep{selfharmony2025}, $\mathcal F_1$ prompts the model to
rewrite the query and solve it within a single generation
(see Appendix~\ref{app:reframe_instruction} for the exact
instruction and one-shot demonstration).
The instruction encourages stylistic variation while requiring
preservation of task-critical entities, answer options, relations,
and numerical values. It prohibits adding descriptions of image
content to avoid introducing erroneous visual interpretations
into the rewritten query. Each reframed rollout independently
generates a rewrite and its solution, conditioned on the
corresponding image.

\subsection{Weighted Harmonic Consensus}
\label{sec:method_whms}

Let $n_v(a)$ count rollouts from view $v$ with extracted answer
$a$, and let $N_v=\sum_a n_v(a)$ count successful extractions.
We form the shared candidate set
$\mathcal A=\bigcup_{v\in\mathcal V}\{a:n_v(a)>0\}$.
For nonempty $\mathcal A$, additive smoothing prevents a zero
count in any single view from vetoing a candidate:
\begin{equation}
    \tilde p_v(a)
    =\frac{n_v(a)+\alpha}{N_v+\alpha|\mathcal A|},
    \qquad a\in\mathcal A,\quad \alpha>0.
    \label{eq:smoothed_view_prob}
\end{equation}
Given weights $w_v>0$ satisfying $\sum_v w_v=1$, we select
the pseudo-label by maximizing the weighted harmonic score:
\begin{equation}
    H_w(a)
    =\left(\sum_{v\in\mathcal V}
        \frac{w_v}{\tilde p_v(a)}\right)^{-1},
    \qquad y^*=\arg\max_{a\in\mathcal A}H_w(a).
    \label{eq:whms_selection}
\end{equation}
The score penalizes weak support in individual views according
to their weights. For fixed view weights, normalizing these
scores yields the unique minimizer of a weighted Pearson
$\chi^2$ consensus objective, in which each weight controls
the penalty for disagreement with the corresponding view.
Section~\ref{sec:whms_theory} establishes this result without
requiring correct-answer invariance and analyzes the resulting
rankings. Uniform weighting over two views recovers
Self-Harmony's selector, up to smoothing
\citep{selfharmony2025}. If $\mathcal A=\varnothing$, all
rollout rewards are zero and the instance is excluded from
reliability estimation.

\subsection{Adaptive Reliability Weighting}
\label{sec:method_adaptive}

We adapt view weights using modality-level shifts in pseudo-label
confidence as a heuristic reliability signal, with batch averaging
and temporal smoothing to reduce sampling fluctuations.

\paragraph{Modality-level confidence shifts.}
Let $p_{it}=\tilde p_{v_{it}}(y^*)$ denote the probability
assigned to the current pseudo-label by view $v_{it}$.
The image and text shifts average the absolute confidence change
from varying one modality across both settings of the other:
\begin{equation}
    d_\rho=\frac{|p_{10}-p_{00}|+|p_{11}-p_{01}|}{2},
    \qquad
    d_\tau=\frac{|p_{01}-p_{00}|+|p_{11}-p_{10}|}{2}.
    \label{eq:modality_shift}
\end{equation}
For $m\in\{\rho,\tau\}$, let $d_m^{(s)}$ denote the average
shift over instances with valid pseudo-labels in minibatch $s$.
We map it to a target reliability factor
$r_m^{(s)}=\exp(-\beta_m d_m^{(s)})$ and apply exponential
smoothing:
\begin{equation}
    \bar r_m^{(s)}
    =(1-\eta)\bar r_m^{(s-1)}+\eta r_m^{(s)},
    \qquad \bar r_m^{(0)}=1,
    \label{eq:reliability_ema}
\end{equation}
where $\beta_m>0$ controls shift sensitivity and $\eta\in(0,1]$
controls the update rate.

\paragraph{Factorized view weights.}
For each modality, the original state receives a fixed factor
of $1$ and the transformed state receives the preceding step's
smoothed factor:
$\boldsymbol{\rho}^{(s)}=(1,\bar r_\rho^{(s-1)})$ and
$\boldsymbol{\tau}^{(s)}=(1,\bar r_\tau^{(s-1)})$.
Their normalized products define the view weights:
\begin{equation}
    w_{it}^{(s)}
    =\frac{\rho_i^{(s)}\tau_t^{(s)}}
    {\sum_{j=0}^{1}\sum_{\ell=0}^{1}
     \rho_j^{(s)}\tau_\ell^{(s)}}.
    \label{eq:factorized_view_weight}
\end{equation}
Initialization yields uniform weights. Lower modality reliability
reduces the relative influence of both views using that
transformation, with both factors applied to the doubly transformed
view. All weights remain positive. Step-$s$ pseudo-labels use
these weights, while current shifts update the factors for
step $s+1$. The rule favors stability of pseudo-label support;
confidence shifts alone cannot distinguish degraded evidence
from beneficial corrections.

\subsection{Confidence-Gated Reframing Rewards and Pooled Optimization}
\label{sec:method_reward}

Following Self-Harmony's reframing-based self-play
\citep{selfharmony2025}, we encourage reframed responses that
agree with the pseudo-label while retaining diversity across
cross-view answer distributions. For an optimization rollout
$k$ from view $v_{it}$, let
$\delta_{it,k}=\mathbb I[\hat a_{it,k}=y^*]$. Direct-solving
responses receive reward $R_{i0,k}=\delta_{i0,k}$, while
reframed responses receive
$R_{i1,k}=\delta_{i1,k}(1+b_i)\phi_{i1,k}$, where
$\phi_{i1,k}$ penalizes reframing format violations and $b_i$ is a
confidence-gated diversity bonus.

Specifically, we compute cross-view diversity for image state
$i$ as
$J_i=\mathrm{JS}(\tilde p_{v_{i0}},\tilde p_{v_{i1}})/\log 2$
and define
$b_i=w_d(1-p_{\mathrm{pool},i})J_i$, where
$p_{\mathrm{pool},i}=(2K)^{-1}\sum_k(\delta_{i0,k}+\delta_{i1,k})$.
The gate reduces diversity rewards when optimization rollouts
already agree with the pseudo-label, preventing diversity bonuses
from assigning negative advantages to matching direct-solving
responses.

To preserve the contribution of the diversity bonus, we jointly
normalize rewards across the $4K$ optimization rollouts:
\begin{equation}
    A_{it,k}=
    \frac{R_{it,k}-\mu_{\mathrm{pool}}}
    {\sigma_{\mathrm{pool}}+\epsilon},
    \qquad
    \mu_{\mathrm{pool}}
    =\frac{1}{4K}
    \sum_{i,t,k}R_{it,k}.
    \label{eq:method_pooled_advantage}
\end{equation}
The resulting advantages are applied to response tokens in GRPO
\citep{shao2024deepseekmath}, including both rewritten-query and
solution tokens for reframed responses. The confidence-gating
property and its guarantee that $\mu_{\mathrm{pool}}\le1$ are
provided in Appendix~\ref{app:reward_analysis}.

\begin{algorithm}[ht]
\caption{\method: Adaptive Harmonic Consensus}
\label{alg:timbre}
\small
\begin{algorithmic}[1]
\Require Policy $\pi_\theta$, unlabeled minibatches
$\{\mathcal B_s\}_{s=1}^{S}$, rollout counts $M>K$
\State Initialize $\bar r_\rho^{(0)}=\bar r_\tau^{(0)}=1$
\For{$s=1,\ldots,S$}
    \State Form weights $w^{(s)}$ from previous reliability
    factors \hfill Eq.~\ref{eq:factorized_view_weight}
    \ForAll{$(I,T)\in\mathcal B_s$}
        \State Construct four views and sample $M$ rollouts
        per view \hfill Eq.~\ref{eq:view_grid}
        \State Estimate $\tilde p_v$ and select $y^*$
        \hfill Eqs.~\ref{eq:smoothed_view_prob}--\ref{eq:whms_selection}
        \State Compute modality shifts $d_\rho,d_\tau$
        \hfill Eq.~\ref{eq:modality_shift}
        \State Compute $J_i$ from all $M$ rollouts and gate
        bonuses using the first $K$
        \hfill Eqs.~\ref{eq:group_jsd}, \ref{eq:confidence_gate}
        \State Assign rewards and pool the $4K$ optimization
        rollouts to obtain advantages
        \hfill Eqs.~\ref{eq:reframer_reward},
        \ref{eq:method_pooled_advantage}
    \EndFor
    \State Update $\pi_\theta$ with GRPO using the pooled advantages
    \State Average valid confidence shifts over $\mathcal B_s$
    and update reliability factors
    \hfill Eq.~\ref{eq:reliability_ema}
\EndFor
\State \Return $\pi_\theta$
\end{algorithmic}
\end{algorithm}
\section{Theoretical Analysis}
\label{sec:whms_theory}

We characterize weighted harmonic consensus through a distributional
optimization objective, establish its exact relation to arithmetic
support, and analyze how modality reweighting affects candidate
rankings. Throughout, we fix an instance with a nonempty candidate
set $\mathcal A$, strictly positive smoothed distributions
$\tilde p_v(a)$, and weights $w_v>0$ satisfying $\sum_v w_v=1$.
When discussing correctness, we assume the correct answer
$C\in\mathcal A$. Proofs and additional results appear in
Appendix~\ref{app:whms_theory}.

\subsection{Harmonic Consensus as a Weighted $\chi^2$ Barycenter}
\label{sec:chi2_barycenter}

We seek a consensus distribution that balances agreement with
individual views according to their assigned reliability weights.
For fixed weights, we formulate this as minimizing a weighted
sum of Pearson $\chi^2$ divergences \citep{pearson1900criterion}:
\begin{equation}
    z^*=\arg\min_{z\in\Delta(\mathcal A)}
    \sum_v w_v D_{\chi^2}(z\|\tilde p_v),
    \qquad
    D_{\chi^2}(z\|\tilde p_v)
    =\sum_{a\in\mathcal A}
    \frac{(z(a)-\tilde p_v(a))^2}{\tilde p_v(a)},
    \label{eq:chi2_barycenter}
\end{equation}
where $\Delta(\mathcal A)$ is the probability simplex.
This divergence penalizes deviations relative to each view's
support, making it costly to assign substantial consensus mass
to candidates that a positively weighted view scarcely supports.
Each weight scales the corresponding disagreement penalty,
controlling how strongly that view constrains the consensus.

\begin{lemma}[Weighted $\chi^2$ barycenter]
\label{lem:chi2_barycenter}
The objective in Eq.~\ref{eq:chi2_barycenter} is strictly convex
and has the unique minimizer
\begin{equation}
    z^*(a)=\frac{H_w(a)}{\sum_{b\in\mathcal A}H_w(b)},
    \qquad
    H_w(a)=
    \left(\sum_v\frac{w_v}{\tilde p_v(a)}\right)^{-1}.
    \label{eq:chi2_solution}
\end{equation}
Consequently, maximizing $H_w$ selects a mode of the barycenter.
\end{lemma}

This result gives an exact variational characterization of our
pseudo-label selector without requiring correct-answer invariance.
Whether the resulting selector favors the correct answer depends
on the relative support and dispersion of the competing
candidates, which we characterize next.

\subsection{Exact Support--Dispersion and Ranking Analysis}
\label{sec:exact_ranking}

For candidate $a$, define its weighted arithmetic support and
dimensionless cross-view dispersion as
\begin{equation}
    m_a=\sum_v w_v\tilde p_v(a),
    \qquad
    \Gamma_w(a)=\frac{1}{m_a}
    \sum_v w_v\frac{(\tilde p_v(a)-m_a)^2}{\tilde p_v(a)}.
    \label{eq:support_dispersion}
\end{equation}

\begin{proposition}[Exact support--dispersion decomposition]
\label{prop:exact_decomposition}
For every $a\in\mathcal A$,
\begin{equation}
    H_w(a)=\frac{m_a}{1+\Gamma_w(a)}\le m_a,
    \label{eq:exact_decomposition}
\end{equation}
with equality if and only if $\tilde p_v(a)$ is constant across views.
\end{proposition}

Harmonic consensus therefore discounts arithmetic support by
dispersion, with low-support views contributing more strongly to
the discount. This discount is itself a $\chi^2$ divergence:
Appendix~\ref{app:whms_theory} shows that
$\Gamma_w(a)=D_{\chi^2}(w\|s_a)$, where
$s_a(v)=w_v\tilde p_v(a)/m_a$ is the share of $a$'s weighted
support contributed by view $v$, so the discount measures how far
the assigned weights are from the support a candidate actually
receives. Whether this improves a ranking depends on the
discounts of both candidates.

\begin{theorem}[Exact pairwise ranking]
\label{thm:exact_pairwise}
For the correct answer $C$ and any competitor
$A\in\mathcal A\setminus\{C\}$,
\begin{equation}
    H_w(C)>H_w(A)
    \iff
    \Gamma_w(A)>
    \frac{m_A}{m_C}\bigl(1+\Gamma_w(C)\bigr)-1.
    \label{eq:exact_pairwise}
\end{equation}
\end{theorem}

This criterion characterizes the ranking behavior of harmonic
aggregation. A correct arithmetic ranking is preserved whenever
$m_C>(1+\Gamma_w(C))m_A$, regardless of the competitor's
dispersion. Exact correct-answer invariance therefore preserves
every strict arithmetic win.
An arithmetic error $m_A>m_C$ is corrected when the
competitor's dispersion exceeds the threshold in
Eq.~\ref{eq:exact_pairwise}. Harmonic aggregation can thus
preserve correct rankings and recover additional ones when
correct-answer support is stable and competing answers vary
across views.

\subsection{Exact Sensitivity to Modality Reweighting}
\label{sec:weight_sensitivity}

Suppressing the update index, let $r=\bar r_\rho$ and
$u=\bar r_\tau$. The factorized weights can be written as
\begin{equation}
    q_\rho=\frac{(1,r)}{1+r},
    \qquad q_\tau=\frac{(1,u)}{1+u},
    \qquad w_{it}=q_\rho(i)q_\tau(t).
    \label{eq:weight_marginals}
\end{equation}
Thus $q_\rho(i)=\sum_t w_{it}$ is the total weight of image
state $i$, and similarly for text. Holding the view distributions
and $q_\tau$ fixed, define
\begin{equation}
    B_i(a)=\sum_{t=0}^{1}
    \frac{q_\tau(t)}{\tilde p_{v_{it}}(a)},
    \qquad h_i(a)=B_i(a)^{-1},
    \qquad H_r(a)=\frac{1+r}{B_0(a)+rB_1(a)}.
    \label{eq:group_reciprocal_support}
\end{equation}
Here $h_i(a)$ is the weighted harmonic score over the two text
modes within image state $i$. Let $D_i(A)=B_i(A)-B_i(C)$.
Then $D_i(A)>0$ precisely when $h_i(C)>h_i(A)$, meaning that
the aggregated evidence within image state $i$ favors $C$
over $A$.

\begin{theorem}[Modality-weight sensitivity]
\label{thm:weight_sensitivity}
For fixed view distributions and text weights, define the
reciprocal ranking margin
\begin{equation}
    \Delta_A(r)=H_r(A)^{-1}-H_r(C)^{-1}
    =\frac{D_0(A)+rD_1(A)}{1+r}.
    \label{eq:reciprocal_margin}
\end{equation}
Then $H_r(C)>H_r(A)$ if and only if $\Delta_A(r)>0$, and
\begin{equation}
    \frac{\partial\Delta_A(r)}{\partial r}
    =\frac{D_1(A)-D_0(A)}{(1+r)^2}.
    \label{eq:weight_derivative}
\end{equation}
Consequently, any decrease in $r$ strictly increases this margin
if and only if $D_0(A)>D_1(A)$. The analogous statement holds
for text reweighting with the modalities exchanged.
\end{theorem}

Since both arithmetic support and dispersion depend on the
weights, this derivative captures their combined effect on
the reciprocal ranking margin.

\paragraph{Implications for adaptive weighting.}
The analysis compares the correct answer $C$ with a competitor
$A$ for a fixed instance. Holding view distributions and text
weights fixed, downweighting the image transformation increases
the reciprocal ranking margin precisely when $D_0(A)>D_1(A)$.
Transformation-induced loss of task-relevant information
provides one source of this imbalance.

Because $C$ is unavailable during adaptation, we use shifts
in support for the current pseudo-label as an observable
stability heuristic. Batch averaging and temporal smoothing
aggregate this signal across instances and updates, reducing
sampling fluctuations and producing shared modality-level
reliability factors. These factors are intended to reflect
the reliability of the model's reframing behavior and the
applied image transformations. The analysis characterizes
when downweighting increases an instance-level pairwise margin;
we provide an empirical analysis of the resulting reliability
dynamics in Appendix~\ref{app:reliability_analysis}, showing
that the estimated reliability factors are aligned with the
empirical per-view pseudo-label accuracy.
\section{Experiments}
\label{sec:experiments}

\subsection{Experimental Setup}
\label{sec:experimental_setup}

\paragraph{Benchmarks and data splits.}
We evaluate \method{} on five visual mathematical reasoning
benchmarks: MathVista \citep{lu2024mathvista}, MathVision
\citep{wang2024mathvision}, MathVerse \citep{zhang2024mathverse},
DynaMath \citep{zou2025dynamath}, and WeMath
\citep{qiao2025wemath}. We also evaluate knowledge-intensive
multimodal reasoning on MMMU-Pro's Science subset
\citep{yue2025mmmupro} in the standard four-option setting.
For each benchmark, we construct an unlabeled adaptation set
and a separate held-out evaluation set to assess adaptation
and within-benchmark transfer. Ground-truth answers are used
only for evaluation. Appendix~\ref{app:data_construction}
details the data construction and split sizes.

\paragraph{Models and implementation.}
We use the 2B, 4B, and 8B models from the Qwen3-VL-Instruct
family \citep{bai2025qwen3vl} and implement \method{} within
the \texttt{verl} framework \citep{sheng2025hybridflow}.
All experiments run on a single node equipped with eight
NVIDIA A100 GPUs. We compare against the unadapted model
(\emph{Before RL}) and TTRV \citep{ttrv2025}. Across adaptation
methods, we match the per-instance budgets of 64 rollouts for
distribution estimation and 32 rollouts for policy optimization.
For our four-view method, each view contributes $M=16$
estimation rollouts, of which the first $K=8$ are used for
optimization, giving $4M=64$ and $4K=32$ in total.
Appendix~\ref{app:hyperparameters} provides the detailed hyperparameters.

\paragraph{Evaluation protocol.}
We report pass@1 (\%), estimated by averaging correctness over
eight sampled responses per instance and then across instances.
\textsc{Val-TTT} evaluates the adapted model on the instances
used for label-free adaptation, measuring target-task performance.
\textsc{Val} evaluates the adapted model on held-out instances
without further parameter updates, measuring within-benchmark
transfer. Each table entry reports
\textsc{Val-TTT} / \textsc{Val}.

\subsection{Main Results}
Table~\ref{tab:main_results} compares \method{} with TTRV and
the unadapted model under matched rollout budgets. Across the
18 benchmark--model combinations, \method{} achieves higher Val-TTT performance in 16 cases, demonstrating improved
label-free adaptation on target instances. For held-out
evaluation, \method{} achieves higher Val performance
in 17 of 18 cases, demonstrating strong transfer without
additional updates. These results suggest that reliability-aware
harmonic cross-view consensus provides more effective learning
signals than single-presentation frequency-based self-reward
construction.

\begin{table*}[h]
    \centering
    \small
    \setlength{\tabcolsep}{2.5pt}
    \renewcommand{\arraystretch}{1.08}
    \caption{Adaptation and held-out performance (pass@1, \%).
    Each entry reports \textbf{Val-TTT / Val}: performance
    on adaptation instances / held-out instances.
    Bold and underline indicate the best and second-best
    distinct scores, respectively, within each model size,
    benchmark, and evaluation setting.}
    \label{tab:main_results}
    \begin{tabular*}{\textwidth}
        {@{\extracolsep{\fill}}l|*{5}{c}|c@{}}
        \toprule
        & \multicolumn{5}{c|}{\textbf{Visual Mathematical Reasoning}}
        & \textbf{Multi-Discipline} \\
        \cmidrule(lr){2-6}\cmidrule(lr){7-7}
        \textbf{Method}
        & MathVista & MathVision & MathVerse & DynaMath & WeMath
        & MMMU-Pro (Science) \\
        \midrule
        \multicolumn{7}{c}{\textit{Qwen3-VL-2B-Instruct}} \\
        \midrule
        Before RL
        & \evalpair{66.3}{63.1}
        & \evalpair{26.9}{\textbf{28.5}}
        & \evalpair{39.8}{42.8}
        & \evalpair{53.3}{53.6}
        & \evalpair{62.3}{60.7}
        & \evalpair{\underline{50.3}}{\textbf{46.4}} \\
        - TTRV
        & \evalpair{\underline{72.8}}{\underline{66.7}}
        & \evalpair{\underline{28.7}}{27.5}
        & \evalpair{\underline{46.7}}{\underline{47.8}}
        & \evalpair{\underline{58.7}}{\underline{57.2}}
        & \evalpair{\underline{71.4}}{\underline{66.3}}
        & \evalpair{\textbf{50.6}}{43.0} \\
        - \method{} (ours)
        & \evalpair{\textbf{74.2}}{\textbf{68.8}}
        & \evalpair{\textbf{29.2}}{\underline{28.3}}
        & \evalpair{\textbf{47.0}}{\textbf{49.4}}
        & \evalpair{\textbf{60.6}}{\textbf{59.1}}
        & \evalpair{\textbf{73.8}}{\textbf{69.6}}
        & \evalpair{\textbf{50.6}}{\underline{45.8}} \\
        \midrule
        \multicolumn{7}{c}{\textit{Qwen3-VL-4B-Instruct}} \\
        \midrule
        Before RL
        & \evalpair{77.0}{74.2}
        & \evalpair{\underline{48.4}}{\textbf{48.8}}
        & \evalpair{55.0}{57.6}
        & \evalpair{65.7}{66.4}
        & \evalpair{79.4}{78.2}
        & \evalpair{61.7}{54.2} \\
        - TTRV
        & \evalpair{\underline{79.7}}{\underline{74.7}}
        & \evalpair{48.2}{47.6}
        & \evalpair{\textbf{56.1}}{\underline{57.5}}
        & \evalpair{\underline{68.4}}{\underline{66.9}}
        & \evalpair{\underline{83.8}}{\underline{82.6}}
        & \evalpair{\underline{64.0}}{\underline{55.0}} \\
        - \method{} (ours)
        & \evalpair{\textbf{81.1}}{\textbf{75.5}}
        & \evalpair{\textbf{49.5}}{\underline{47.9}}
        & \evalpair{\underline{55.8}}{\textbf{58.0}}
        & \evalpair{\textbf{69.2}}{\textbf{67.7}}
        & \evalpair{\textbf{84.8}}{\textbf{83.1}}
        & \evalpair{\textbf{64.5}}{\textbf{55.4}} \\
        \midrule
        \multicolumn{7}{c}{\textit{Qwen3-VL-8B-Instruct}} \\
        \midrule
        Before RL
        & \evalpair{78.4}{74.5}
        & \evalpair{51.8}{\textbf{48.8}}
        & \evalpair{55.0}{60.3}
        & \evalpair{67.0}{68.0}
        & \evalpair{80.2}{79.5}
        & \evalpair{64.0}{\textbf{58.7}} \\
        - TTRV
        & \evalpair{\underline{81.6}}{\underline{76.0}}
        & \evalpair{\underline{52.4}}{47.6}
        & \evalpair{\underline{56.9}}{\underline{61.0}}
        & \evalpair{\underline{68.4}}{\underline{68.1}}
        & \evalpair{\underline{84.8}}{\underline{83.2}}
        & \evalpair{\underline{65.9}}{\underline{57.8}} \\
        - \method{} (ours)
        & \evalpair{\textbf{82.5}}{\textbf{76.5}}
        & \evalpair{\textbf{52.9}}{\underline{47.9}}
        & \evalpair{\textbf{57.3}}{\textbf{62.4}}
        & \evalpair{\textbf{69.7}}{\textbf{69.1}}
        & \evalpair{\textbf{86.0}}{\textbf{85.2}}
        & \evalpair{\textbf{66.6}}{57.0} \\
        \bottomrule
    \end{tabular*}
\end{table*}

\begin{figure}[t]
    \centering
    \includegraphics[width=0.9\linewidth]{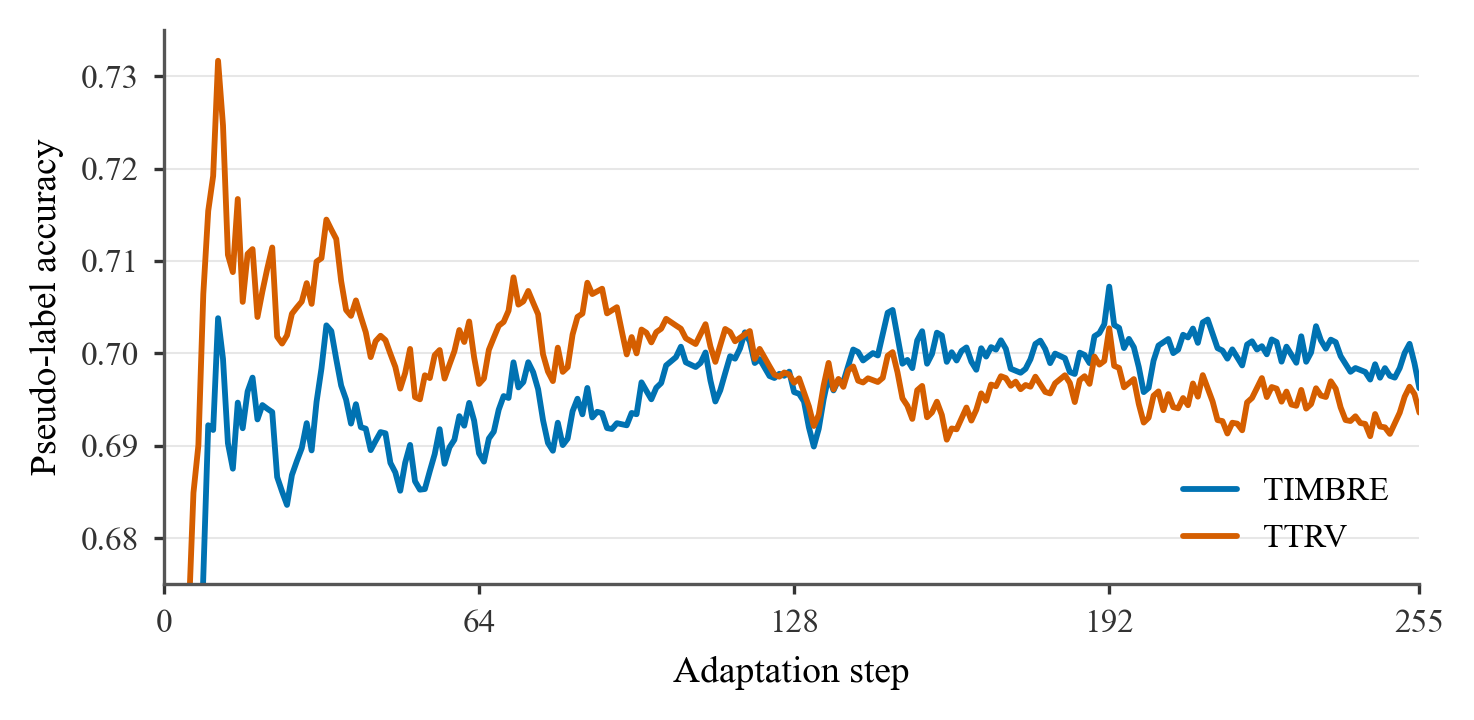}
    \caption{
    \textbf{Pseudo-label accuracy during test-time adaptation.}
    Comparison between \method{} and TTRV on DynaMath with
    Qwen3-VL-8B-Instruct. Curves are smoothed using a time-weighted
    exponential moving average for visualization. Ground-truth answers
    are used only for post-hoc evaluation and are unavailable during
    adaptation. \method{} maintains higher pseudo-label accuracy over
    most later adaptation steps, indicating improved reliability of the
    generated training signal.
    }
    \label{fig:label_accuracy}
\end{figure}

We further analyze the quality of pseudo-labels generated during
test-time adaptation. Specifically, we compare the pseudo-label
accuracy of \method{} and TTRV on DynaMath with Qwen3-VL-8B-Instruct.
Although ground-truth answers are unavailable during adaptation,
we use them only for post-hoc evaluation of the selected pseudo-labels.
As shown in Figure~\ref{fig:label_accuracy}, \method{} produces more
reliable pseudo-labels throughout adaptation, maintaining consistently
higher label accuracy during later adaptation steps. This indicates that
weighted harmonic cross-view consensus provides a more stable and
effective supervision signal than single-presentation
frequency-based pseudo-label selection.

\subsection{Ablation of Consensus Aggregation}

To isolate the contribution of the weighted harmonic consensus,
we compare three aggregation strategies:
(1) weighted harmonic aggregation used by \method{},
(2) weighted arithmetic aggregation using the same adaptive reliability weights, and
(3) uniform harmonic aggregation following Self-Harmony.
The weighted arithmetic variant preserves explicit view
reliability weighting but removes the dispersion discount
introduced by harmonic aggregation. Uniform harmonic aggregation
retains harmonic consensus but assigns equal influence to all
views. All variants use the same
four-view generation procedure and optimization settings.
Results on Qwen3-VL-4B over MathVista and DynaMath are shown
in Table~\ref{tab:aggregation_ablation}.
The comparison evaluates the contribution of both reliability-aware
weighting and harmonic dispersion discounting.

\begin{table}[h]
    \centering
    \small
    \setlength{\tabcolsep}{4pt}
    \renewcommand{\arraystretch}{1.1}
    \caption{Ablation of consensus aggregation strategies on
    Qwen3-VL-4B. Each entry reports \textsc{Val-TTT} /
    \textsc{Val} pass@1 (\%). All variants use the same
    optimization settings.}
    \label{tab:aggregation_ablation}
    \begin{tabular}{lcc}
        \toprule
        \textbf{Method}
        & MathVista
        & DynaMath \\
        \midrule
        Before RL
        & \evalpair{77.0}{74.2}
        & \evalpair{65.7}{66.4} \\

        - Weighted arithmetic aggregation
        & \evalpair{80.1}{75.4}
        & \evalpair{68.2}{67.6} \\

        - Uniform harmonic aggregation
        & \evalpair{79.5}{75.0}
        & \evalpair{\textbf{69.2}}{67.5} \\

        - Weighted harmonic aggregation (\method{})
        & \evalpair{\textbf{81.1}}{\textbf{75.5}}
        & \evalpair{\textbf{69.2}}{\textbf{67.7}} \\
        \bottomrule
    \end{tabular}
\end{table}
\section{Conclusion}
\label{sec:conclusion}

We introduced \textsc{TIMBRE}, a reliability-aware harmonic consensus framework
for label-free multimodal test-time reinforcement learning. \textsc{TIMBRE}
selects pseudo-labels through weighted harmonic aggregation over four
image--text views and adapts modality-level weights online using a
label-free reliability heuristic. Without requiring correct-answer
invariance, we showed that the normalized weighted harmonic aggregate is
the unique minimizer of a weighted Pearson $\chi^2$-barycenter objective,
and characterized its ranking behavior under cross-view dispersion and
modality reweighting. Under matched rollout budgets, \textsc{TIMBRE}
improves over TTRV in 16 of 18 benchmark--model settings on adaptation
instances and 17 of 18 settings on held-out instances. Together, these
results support reliability-aware harmonic consensus as an effective
approach to improving self-generated supervision for label-free multimodal
test-time reinforcement learning.

\newpage

\section*{AI use statement}

In this work, we used generative AI tools for assistance with software code
during implementation and debugging. We have not used generative AI tools for
developing theoretical models or conceptual frameworks, formulating mathematical claims, providing ingredients for or writing proofs, proposing or refining hypotheses, designing or providing feedback on research methodology or
experiments, translation, cleaning or reformatting datasets, or interpreting
results. Generating synthetic datasets and qualitative or thematic data analysis are not applicable to this work.

Additionally, we used generative AI tools for editing and polishing the writing
of the paper to improve clarity and readability, and for retrieval and discovery, such as searching for and identifying related work. We have reviewed all AI-assisted work. AI-assisted code was inspected and tested by the authors before being used in the reported experiments; all edited text was checked to ensure that it accurately reflects our methods, theory, and results; and all references identified with AI assistance were verified against the original sources before being cited. We take responsibility for the final content of this work, including text, claims or artifacts produced with the aid of generative AI.

\section*{Ethics statement}

This work uses only publicly available benchmarks (MathVista, MathVision,
MathVerse, DynaMath, WeMath, and MMMU-Pro) and publicly released models from
the Qwen3-VL family, in accordance with their respective licenses and terms of
use. It involves no human subjects, crowdsourcing, or collection of personal
or sensitive data. Because label-free test-time reinforcement learning derives
rewards from a model's own predictions, it can reinforce incorrect consensus
when pseudo-labels are wrong; our method is designed to mitigate this failure
mode but does not eliminate it, so adapted models should be validated before
deployment in high-stakes settings.

\section*{Reproducibility statement}

We describe the full TIMBRE procedure in Section~\ref{sec:method} and Algorithm~\ref{alg:timbre}, including
view construction (Section~
\ref{sec:method_views}), weighted harmonic consensus (Section~\ref{sec:method_whms}),
adaptive reliability weighting (Section~\ref{sec:method_adaptive}), and reward construction with
pooled optimization (Section~\ref{sec:method_reward}). All theoretical results state their
assumptions in Section~\ref{sec:whms_theory}, with complete proofs in Appendix~\ref{app:whms_theory}. Data sources,
configurations, and adaptation/held-out splits for all six benchmarks are
given in Section~\ref{sec:experimental_setup} and Appendix~\ref{app:data_construction}; hyperparameters and rollout
budgets are listed in Section~\ref{sec:experimental_setup} and Appendix~\ref{app:hyperparameters}.

\bibliography{iclr2027_conference}
\bibliographystyle{iclr2027_conference}

\newpage

\appendix

\section{Additional Theory for Weighted Harmonic Consensus}
\label{app:whms_theory}

We provide proofs of the barycenter characterization, exact ranking
results, and modality-weight sensitivity, followed by boundary cases,
the local variance approximation, and the scope of the reliability
heuristic. All results are conditional on the fixed, positive
estimated distributions specified in Section~\ref{sec:whms_theory}.

\subsection{Proof of the Weighted $\chi^2$ Barycenter Characterization}
\label{app:chi2_barycenter}

\begin{proof}[Proof of Lemma~\ref{lem:chi2_barycenter}]
Let $S_a=\sum_v w_v/\tilde p_v(a)=H_w(a)^{-1}>0$.
Using $\sum_a z(a)=\sum_a\tilde p_v(a)=1$ and
$\sum_v w_v=1$, we obtain
\begin{equation}
\begin{aligned}
    \sum_v w_vD_{\chi^2}(z\|\tilde p_v)
    &=
    \sum_a\left[
        S_a z(a)^2-2z(a)+\sum_v w_v\tilde p_v(a)
    \right]\\
    &=\sum_a S_a z(a)^2-1.
\end{aligned}
\label{eq:app_expanded_chi2}
\end{equation}
The objective is strictly convex because its Hessian is
$\operatorname{diag}(2S_a)\succ0$. Applying Cauchy--Schwarz
to $\{\sqrt{S_a}z(a)\}_a$ and $\{S_a^{-1/2}\}_a$ gives
\begin{equation}
    1=\left(\sum_a z(a)\right)^2
    \le
    \left(\sum_a S_a z(a)^2\right)
    \left(\sum_a S_a^{-1}\right).
    \label{eq:app_barycenter_bound}
\end{equation}
Equality holds exactly when
$\sqrt{S_a}z(a)=cS_a^{-1/2}$ for a common constant $c$,
or equivalently $z(a)=cS_a^{-1}$.
Normalization therefore yields
\begin{equation}
    z^*(a)
    =\frac{S_a^{-1}}{\sum_b S_b^{-1}}
    =\frac{H_w(a)}{\sum_b H_w(b)}.
    \label{eq:app_chi2_solution}
\end{equation}
This distribution is strictly positive and attains the lower
bound, so it is the unique minimizer. Its shared positive
normalizer preserves all rankings and ties.
\end{proof}

The minimum objective value is
$(\sum_a H_w(a))^{-1}-1$.
The weighted AM--HM inequality gives
$H_w(a)\le\sum_v w_v\tilde p_v(a)$ and hence
$\sum_aH_w(a)\le1$, with equality exactly when all view
distributions coincide. Thus, the barycenter balances disagreement
without requiring identical views. This characterization depends
on the divergence direction: reversing its arguments does not
generally yield normalized harmonic scores.

\subsection{Proof of the Support--Dispersion Decomposition}
\label{app:exact_decomposition}

\begin{proof}[Proof of Proposition~\ref{prop:exact_decomposition}]
Define
$V_w^\chi(a)=
\sum_v w_v(\tilde p_v(a)-m_a)^2/\tilde p_v(a)$.
Expanding gives
\begin{equation}
\begin{aligned}
    V_w^\chi(a)
    &=\sum_v w_v\left[
        \tilde p_v(a)-2m_a+\frac{m_a^2}{\tilde p_v(a)}
    \right]\\
    &=m_a^2H_w(a)^{-1}-m_a.
\end{aligned}
\label{eq:app_vchi_identity}
\end{equation}
Dividing by $m_a>0$ gives
$\Gamma_w(a)=m_aH_w(a)^{-1}-1$, proving
Eq.~\ref{eq:exact_decomposition}.
\end{proof}

The dispersion also admits a divergence interpretation.
Define $s_a(v)=w_v\tilde p_v(a)/m_a$, the fraction of
candidate $a$'s weighted arithmetic support contributed
by view $v$. Then
\begin{equation}
    D_{\chi^2}(w\|s_a)
    =\sum_v\frac{(w_v-s_a(v))^2}{s_a(v)}
    =\Gamma_w(a).
    \label{eq:app_gamma_chi2}
\end{equation}
Thus, the harmonic discount measures the discrepancy between
the assigned view weights and the candidate-specific support shares.

\subsection{Pairwise Rankings and Invariance}
\label{app:pairwise_analysis}

\begin{proof}[Proof of Theorem~\ref{thm:exact_pairwise}]
By the support--dispersion decomposition and positivity of
all denominators,
\begin{equation}
\begin{aligned}
    H_w(C)>H_w(A)
    &\iff
    m_C(1+\Gamma_w(A))>m_A(1+\Gamma_w(C))\\
    &\iff
    \Gamma_w(A)>
    \frac{m_A}{m_C}(1+\Gamma_w(C))-1.
\end{aligned}
\end{equation}
\end{proof}

\paragraph{Preserving correct arithmetic rankings.}
Since $\Gamma_w(A)\ge0$, the condition
$m_C>(1+\Gamma_w(C))m_A$ guarantees $H_w(C)>H_w(A)$
regardless of the competitor's dispersion.
A sufficient condition follows from relative support stability:
if $|\tilde p_v(C)-m_C|\le\varepsilon m_C$ for every view,
with $0\le\varepsilon<1$, then
$\tilde p_v(C)\ge(1-\varepsilon)m_C$ and
\begin{equation}
    \Gamma_w(C)
    \le
    \frac{\sum_vw_v(\tilde p_v(C)-m_C)^2}
         {m_C^2(1-\varepsilon)}
    \le \frac{\varepsilon^2}{1-\varepsilon}.
    \label{eq:app_relative_invariance}
\end{equation}
Consequently,
\begin{equation}
    m_C>
    \left(1+\frac{\varepsilon^2}{1-\varepsilon}\right)m_A
    \quad\Longrightarrow\quad H_w(C)>H_w(A).
    \label{eq:app_robust_ranking}
\end{equation}
Equivalently, a relative arithmetic margin
$m_C/m_A-1>\varepsilon^2/(1-\varepsilon)$ suffices for
preservation. This sufficient threshold is $O(\varepsilon^2)$
as $\varepsilon\to0$.
Exact invariance corresponds to $\varepsilon=0$ and
preserves every strict arithmetic win.

\paragraph{Correcting arithmetic errors.}
When $m_A>m_C$, harmonic aggregation corrects the ranking
exactly when
$\Gamma_w(A)>(m_A/m_C)(1+\Gamma_w(C))-1$.
Under exact correct-answer invariance, this condition reduces
to $\Gamma_w(A)>m_A/m_C-1$.
Thus, exact invariance guarantees preservation of all strict
arithmetic wins and permits additional wins against sufficiently
dispersed competitors. Under approximate invariance, the
margin condition in Eq.~\ref{eq:app_robust_ranking}
guarantees preservation.

\subsection{Modality Marginals and Exact Weight Sensitivity}
\label{app:weight_sensitivity}

For $r=\bar r_\rho$ and $u=\bar r_\tau$, the normalizer of
the factorized view weights is $(1+r)(1+u)$. Hence
\begin{equation}
    w_{00}+w_{01}=\frac{1}{1+r},
    \qquad
    w_{10}+w_{11}=\frac{r}{1+r},
    \label{eq:app_image_marginals}
\end{equation}
and the corresponding text marginals are $1/(1+u)$ and
$u/(1+u)$. This establishes Eq.~\ref{eq:weight_marginals}.

\begin{proof}[Proof of Theorem~\ref{thm:weight_sensitivity}]
Holding $q_\tau$ and the view distributions fixed gives
\begin{equation}
    H_r(a)^{-1}
    =\sum_i q_\rho(i)\sum_t
    \frac{q_\tau(t)}{\tilde p_{v_{it}}(a)}
    =\frac{B_0(a)+rB_1(a)}{1+r}.
\end{equation}
Subtracting the reciprocal score of $C$ from that of $A$
yields Eq.~\ref{eq:reciprocal_margin}. Since the harmonic
scores are positive,
$\Delta_A(r)>0\iff H_r(C)>H_r(A)$.
Differentiation gives
\begin{equation}
    \Delta_A'(r)
    =\frac{D_1(A)(1+r)-D_0(A)-rD_1(A)}{(1+r)^2}
    =\frac{D_1(A)-D_0(A)}{(1+r)^2}.
\end{equation}
Since $(1+r)^2>0$, the derivative has a constant sign
determined by $D_1(A)-D_0(A)$. Thus, decreasing $r$ strictly
increases $\Delta_A(r)$ if and only if $D_0(A)>D_1(A)$,
and leaves it unchanged when $D_0(A)=D_1(A)$.
Exchanging image and text establishes the analogous result
for text reweighting.
\end{proof}

\begin{corollary}[A sufficient condition for downweighting]
\label{cor:degraded_transform}
If $\tilde p_{v_{1t}}(C)\le\tilde p_{v_{0t}}(C)$ and
$\tilde p_{v_{1t}}(A)\ge\tilde p_{v_{0t}}(A)$
for both $t\in\{0,1\}$, then $D_1(A)\le D_0(A)$.
With view distributions and text weights fixed, any decrease
in $r$ therefore weakly increases $\Delta_A(r)$, with a strict
increase if any of these probability inequalities is strict.
\end{corollary}

\begin{proof}
Taking reciprocals gives
\begin{equation}
    \tilde p_{v_{1t}}(A)^{-1}
    -\tilde p_{v_{1t}}(C)^{-1}
    \le
    \tilde p_{v_{0t}}(A)^{-1}
    -\tilde p_{v_{0t}}(C)^{-1}.
\end{equation}
Averaging over $t$ with strictly positive weights $q_\tau(t)$
yields $D_1(A)\le D_0(A)$, with strict inequality if any
of the assumed probability inequalities is strict.
Theorem~\ref{thm:weight_sensitivity} then establishes
the effect of decreasing $r$.
\end{proof}

\subsection{Reward Construction and Confidence-Gating Guarantee}
\label{app:reward_analysis}

We detail the reframing format penalty, explain the role of
pooled normalization, and establish the confidence-gating
guarantee. Throughout, we consider an instance with a valid
pseudo-label $y^*$.

\paragraph{Reward construction and format penalty.}
For optimization rollout $k\in\{1,\ldots,K\}$ from view
$v_{it}$, define
$\delta_{it,k}=\mathbb I[\hat a_{it,k}=y^*]$,
treating failed answer extractions as nonmatches.
The rewards are
\begin{equation}
    R_{i0,k}=\delta_{i0,k},
    \qquad
    R_{i1,k}=\delta_{i1,k}(1+b_i)\phi_{i1,k}.
    \label{eq:reframer_reward}
\end{equation}
The format multiplier $\phi_{i1,k}$ equals $1$ when a
reframed response follows the required \textsc{strategy},
\textsc{rewritten problem}, and \textsc{solution} structure,
and $w_f\in(0,1)$ otherwise. Thus, positive reward always
requires pseudo-label agreement, while format violations
reduce the reward of matching reframed responses.

For image state $i$, cross-view diversity is measured by
\begin{equation}
    J_i=
    \frac{\mathrm{JS}
    (\tilde p_{v_{i0}},\tilde p_{v_{i1}})}{\log 2}
    \in[0,1],
    \label{eq:group_jsd}
\end{equation}
using the smoothed answer distributions estimated from all
$M$ rollouts per view. The confidence-gated bonus is
\begin{equation}
    b_i=w_d(1-p_{\mathrm{pool},i})J_i,
    \qquad
    p_{\mathrm{pool},i}
    =\frac{1}{2K}\sum_{k=1}^{K}
    (\delta_{i0,k}+\delta_{i1,k}),
    \label{eq:confidence_gate}
\end{equation}
where $0\le w_d\le2$. Unlike $J_i$, the agreement rate
$p_{\mathrm{pool},i}$ uses only the optimization rollouts,
matching the sample pool used to compute advantages.

\paragraph{Why pool rewards across views?}
Separate per-view standardization cancels any shared positive
reward multiplier $c$, apart from numerical stabilization:
\begin{equation}
    \frac{cR-c\mu_v}{c\sigma_v+\epsilon}
    =
    \frac{R-\mu_v}{\sigma_v+\epsilon/c}.
\end{equation}
Within each reframed view, the diversity factor $1+b_i$
is shared across rollouts. Per-view normalization would
therefore largely remove its effect. We instead jointly
standardize all $4K$ optimization rewards for each instance,
as in Eq.~\ref{eq:method_pooled_advantage}, preserving the
bonus's influence on relative rewards across views.

Pooling also motivates confidence gating. With an ungated
bonus $b_i=w_dJ_i$, the pooled mean can exceed $1$, giving
negative advantages to direct-solving responses that match
$y^*$. The gate removes the bonus when all optimization
rollouts within an image state agree with $y^*$ and provides
the following guarantee.

\begin{proposition}[Confidence-gating guarantee]
\label{prop:confidence_gate}
Suppose $0\le w_d\le2$ and $0<\phi_{i1,k}\le1$.
Under Eqs.~\ref{eq:reframer_reward} and
\ref{eq:confidence_gate}, the pooled reward mean satisfies
$\mu_{\mathrm{pool}}\le1$. Consequently, direct-solving
responses that match $y^*$ and format-compliant reframed
responses that match $y^*$ have nonnegative pooled advantages.
\end{proposition}

\begin{proof}
Define the optimization-rollout agreement rate in each view as
$q_{it}=K^{-1}\sum_{k=1}^{K}\delta_{it,k}$.
Then $q_{i0}+q_{i1}=2p_{\mathrm{pool},i}$.
Since $\phi_{i1,k}\le1$,
\begin{equation}
\begin{aligned}
    4\mu_{\mathrm{pool}}
    &=\frac{1}{K}
      \sum_{i=0}^{1}\sum_{k=1}^{K}
      \left[
      \delta_{i0,k}
      +\delta_{i1,k}(1+b_i)\phi_{i1,k}
      \right]\\
    &\le\sum_{i=0}^{1}
      \left[q_{i0}+q_{i1}(1+b_i)\right]\\
    &=\sum_{i=0}^{1}
      \left[
      2p_{\mathrm{pool},i}
      +w_d(1-p_{\mathrm{pool},i})J_iq_{i1}
      \right]\\
    &\le\sum_{i=0}^{1}
      \left[
      2p_{\mathrm{pool},i}
      +2(1-p_{\mathrm{pool},i})
      \right]
      =4,
\end{aligned}
\label{eq:gated_mean_bound}
\end{equation}
where the last inequality uses $w_d\le2$,
$J_i,q_{i1}\le1$, and $p_{\mathrm{pool},i}\in[0,1]$.
Hence $\mu_{\mathrm{pool}}\le1$.

A matching direct-solving response receives reward $1$,
while a matching, format-compliant reframed response receives
$1+b_i\ge1$. Subtracting $\mu_{\mathrm{pool}}\le1$ and
dividing by $\sigma_{\mathrm{pool}}+\epsilon>0$ therefore
gives a nonnegative advantage in both cases.
\end{proof}
\section{Additional Experimental Details and Results}
\label{app:experiments}

\subsection{Reframing Instruction}
\label{app:reframe_instruction}

\paragraph{User message: instruction and demonstration.}
\begin{ReframePrompt}
You are the Reframer, an AI that solves problems with creation.
Your challenge is not just to find the answer, but to find a more interesting way to get there.

Mission:
1. Transform: DO NOT solve the ORIGINAL PROBLEM directly. Rewrite it using one of the strategies below.
2. Solve: Let's think step by step to solve the new problem you created, then output the final answer within \textbackslash{}boxed\{\}.

Strategies (default to Contextualize; use Paraphrase only when no faithful scenario is possible):
- Contextualize: Wrap the question in a concrete real-world situation \textrm{\textemdash} who is looking at this image, and why they need the answer \textrm{\textemdash} then ask the original question from inside that situation. The scenario supplies the asker and the motive; the image supplies everything visual.
- Paraphrase: Restate the question in your own words, with different sentence structure and vocabulary.

Rules:
- Begin your response immediately with "STRATEGY:". Output nothing before it \textrm{\textemdash} no preamble, acknowledgement, or reasoning.
- Do NOT describe the image. Point at it ("the figure", "the chart", "the plants shown") and let the reader look. Never write down identities, colors, counts, positions, shapes, layout, axis values, or labels that you read off the image \textrm{\textemdash} not even ones you are confident about.
- Every fact in the rewritten problem must be traceable to the ORIGINAL PROBLEM TEXT. Facts stated there may be reworded and kept; facts you got by looking belong only in your SOLUTION, never in the problem statement.
- Carry over every quantity, condition, and constraint from the original text. Add none, drop none, and never make a vague condition more specific.
- The rewritten problem must still require the image to solve, and must be self-contained given the image \textrm{\textemdash} solvable from your wording plus the image alone.
- The ORIGINAL PROBLEM is either open-ended (a free-form or numeric answer) or multiple-choice. If it is open-ended, rewrite the question freely. If it is multiple-choice, rewrite ONLY the question stem, then append the original choices verbatim (same letters, same order, same text) so the answer is still one of those choices.
- The final answer must equal the answer to the ORIGINAL PROBLEM.

Response Format (must follow exactly):
STRATEGY: [Chosen strategy]
REWRITTEN PROBLEM:
[Your transformed problem statement - new framing, no description of the image]
SOLUTION:
[Step-by-step solution matching the transformed problem and grounded in image evidence]
\textbackslash{}boxed\{[Final answer]\}

Here is a one-shot reframing example. It shows only the transform step (STRATEGY + REWRITTEN PROBLEM); your own response must still continue on to give the SOLUTION and \textbackslash{}boxed\{\} final answer.

EXAMPLE ORIGINAL PROBLEM:
\end{ReframePrompt}

\begin{center}
    \includegraphics[width=0.65\linewidth]
        {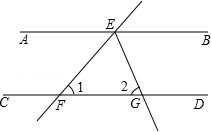}
\end{center}

\begin{ReframePrompt}
As shown in the figure, AB parallel CD, EG bisects angle BEF, angle 1 = 50.0, then angle 2 is equal to ()

EXAMPLE REFRAMING:
STRATEGY: Contextualize
REWRITTEN PROBLEM:
A survey crew is preparing to pave the layout shown in the figure. Roads AB and CD were built parallel, and the crew cut service path EG so that it bisects angle BEF. Their instrument reads angle 1 as 50\textdegree{}. What should they record for angle 2?
\end{ReframePrompt}

\paragraph{Assistant message: fixed acknowledgement.}
\begin{ReframePrompt}
Understood. Please share the original problem.
\end{ReframePrompt}

The original problem message(s), including the corresponding
image(s), follow unchanged. The model then generates the
reframing and solution in a single response.

\newpage
\subsection{Data Construction}
\label{app:data_construction}

We construct subsets for label-free adaptation and held-out
evaluation. Table~\ref{tab:benchmark_splits} summarizes the
data sources and split sizes. Ground-truth answers are used
only for evaluation.

\paragraph{Visual mathematical reasoning benchmarks.}
For MathVista, we randomly partition \texttt{testmini} into
500 adaptation and 500 held-out instances. For MathVerse,
we use the \emph{Text Lite} version of \texttt{testmini},
randomly selecting 500 instances for adaptation and reserving
the remaining 278 for held-out evaluation. For MathVision,
we use the 304 instances in \texttt{testmini} for held-out
evaluation and randomly sample 500 adaptation instances from
\texttt{test}. For DynaMath, we use variants~1 and~2 of
465 seed problems for adaptation and held-out evaluation,
respectively, yielding distinct instances generated from
the same seed problems. For WeMath, we randomly select
500 adaptation and 500 held-out instances without overlap
from the 1,730 instances in \texttt{testmini} to limit
computational cost.

\paragraph{MMMU-Pro.}
We use the Science subset of the \texttt{standard (4 options)}
configuration, covering Biology, Chemistry, Geography,
Mathematics, and Physics.

Although this configuration is distributed as a single
\texttt{test} split, instance identifiers retain their
original MMMU split through the prefixes \texttt{test\_}
and \texttt{validation\_}. We assign instances with the
former prefix to adaptation and those with the latter
to held-out evaluation, yielding 198 adaptation and
93 held-out instances.

Across all benchmarks, adaptation and held-out sets contain
no identical instances.

\begin{table*}[ht]
    \centering
    \small
    \setlength{\tabcolsep}{5pt}
    \caption{Data sources and split sizes for label-free
    adaptation and held-out evaluation.}
    \label{tab:benchmark_splits}
    \begin{tabularx}{\textwidth}{lXrrr}
        \toprule
        Benchmark & Hugging Face source and configuration
        & Adaptation & Held-out & Total \\
        \midrule
        MathVista
        & \texttt{AI4Math/MathVista}
          (\texttt{testmini})
        & 500 & 500 & 1,000 \\

        MathVerse
        & \texttt{AI4Math/MathVerse}
          (\texttt{testmini}, Text Lite)
        & 500 & 278 & 778 \\

        MathVision
        & \texttt{MathLLMs/MathVision}
          (adaptation: \texttt{test};
          held-out: \texttt{testmini})
        & 500 & 304 & 804 \\

        DynaMath
        & \texttt{DynaMath/DynaMath\_Sample}
          (adaptation: variant~1;
          held-out: variant~2)
        & 465 & 465 & 930 \\

        WeMath
        & \texttt{We-Math/We-Math}
          (\texttt{testmini})
        & 500 & 500 & 1,000 \\

        MMMU-Pro (Science)
        & \texttt{MMMU/MMMU\_Pro}
          (standard four-option, Science)
        & 198 & 93 & 291 \\
        \midrule
        \textbf{Total}
        &
        & \textbf{2,663} & \textbf{2,140} & \textbf{4,803} \\
        \bottomrule
    \end{tabularx}
\end{table*}

\newpage
\subsection{Hyperparameters}
\label{app:hyperparameters}

Table~\ref{tab:hyperparameters} summarizes the experimental
hyperparameters. The four-view method allocates the total
estimation and optimization budgets equally across views.
The learning rate remains constant after linear warmup
over the first 3\% of training steps.

\begin{table}[ht]
    \centering
    \caption{Hyperparameters used in our experiments.
    $M$ and $K$ denote per-view rollout counts for \method{}.}
    \label{tab:hyperparameters}
    \scriptsize
    \setlength{\tabcolsep}{3.5pt}
    \renewcommand{\arraystretch}{1.08}
    \begin{tabularx}{\columnwidth}{
        >{\raggedright\arraybackslash}p{0.60\columnwidth}
        >{\raggedright\arraybackslash}X}
        \toprule
        \textbf{Hyperparameter} & \textbf{Value} \\
        \midrule
        \multicolumn{2}{l}{\textit{Training and optimization}} \\
        \midrule
        Maximum prompt length & 6,320 tokens \\
        Maximum response length & 8,192 tokens \\
        Training batch size & 32 \\
        Training epochs & 16 \\
        PPO mini-batch size & 4 \\
        PPO micro-batch size & 2 \\
        Optimizer & Adam \\
        Learning rate & $5\times10^{-7}$ \\
        Learning-rate schedule & Constant after warmup \\
        Warmup ratio & 0.03 \\
        \midrule
        \multicolumn{2}{l}{\textit{Rollout generation and evaluation}} \\
        \midrule
        Total estimation rollouts per instance & 64 \\
        Total optimization rollouts per instance & 32 \\
        Estimation rollouts per view ($M$) & 16 \\
        Optimization rollouts per view ($K$) & 8 \\
        Training temperature & 1.0 \\
        Training top-$p$ & 1.0 \\
        Training top-$k$ & $-1$ (disabled) \\
        Evaluation rollouts per instance & 8 \\
        Evaluation temperature & 0.8 \\
        Evaluation top-$p$ & 0.9 \\
        Evaluation top-$k$ & $-1$ (disabled) \\
        \midrule
        \multicolumn{2}{l}{\textit{Adaptive weighting and rewards}} \\
        \midrule
        Image-shift sensitivity ($\beta_\rho$) & 1.0 \\
        Text-shift sensitivity ($\beta_\tau$) & 1.0 \\
        Reliability EMA update rate ($\eta$) & 0.05 \\
        Initial reliability factors
        ($\bar r_\rho^{(0)},\bar r_\tau^{(0)}$) & $(1,1)$ \\
        Reframing format multiplier ($w_f$) & 0.9 \\
        Diversity bonus coefficient ($w_d$) & 0.5 \\
        \bottomrule
    \end{tabularx}
\end{table}

\subsection{Analysis of Adaptive Reliability Weighting}
\label{app:reliability_analysis}

Because the correct answer $C$ is unavailable during adaptation, our
reliability update cannot directly measure whether a transformation
improves or degrades correctness. Instead, \method{} estimates view
reliability using shifts in support for the current pseudo-label as a
label-free stability heuristic. This appendix analyzes whether the
resulting reliability-aware weights align with the empirical reliability
of different views.

\begin{figure*}[t]
    \centering
    \setlength{\tabcolsep}{2pt}

    \begin{tabular}{cccc}
        \multicolumn{4}{c}{\textbf{Adaptive view weights}} \\[0.2em]

        \includegraphics[width=0.235\textwidth]{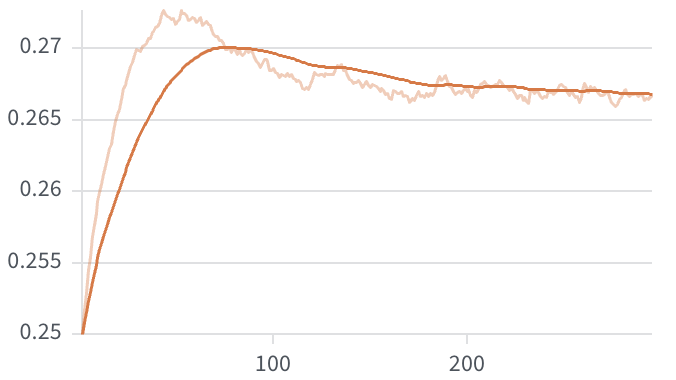} &
        \includegraphics[width=0.235\textwidth]{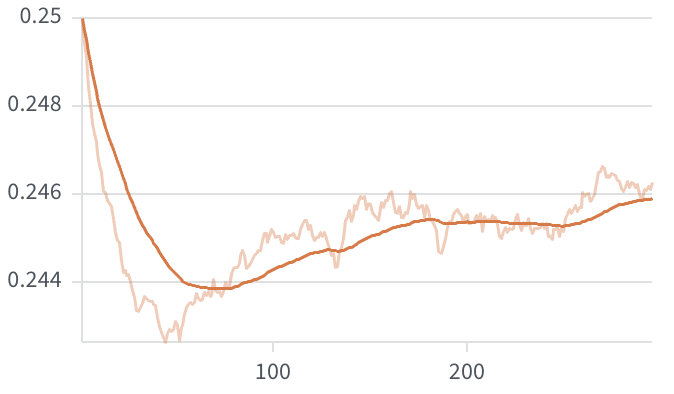} &
        \includegraphics[width=0.235\textwidth]{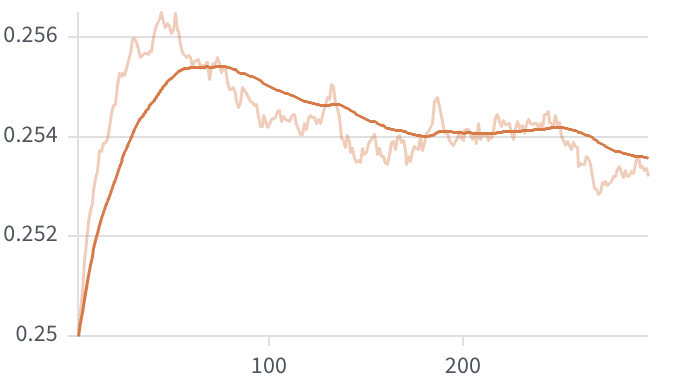} &
        \includegraphics[width=0.235\textwidth]{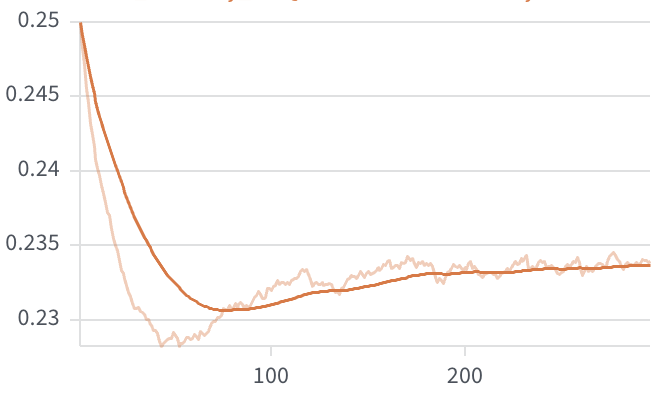}
        \\[-0.3em]

        $w_{00}$ &
        $w_{01}$ &
        $w_{10}$ &
        $w_{11}$ 
        \\[0.5em]

        \multicolumn{4}{c}{\textbf{Per-view pseudo-label accuracy}} \\[0.2em]

        \includegraphics[width=0.235\textwidth]{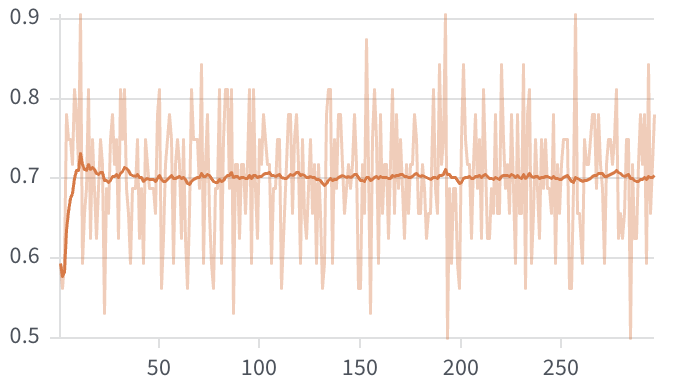} &
        \includegraphics[width=0.235\textwidth]{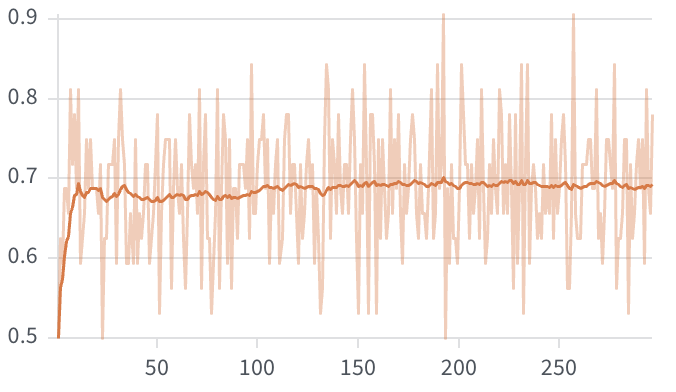} &
        \includegraphics[width=0.235\textwidth]{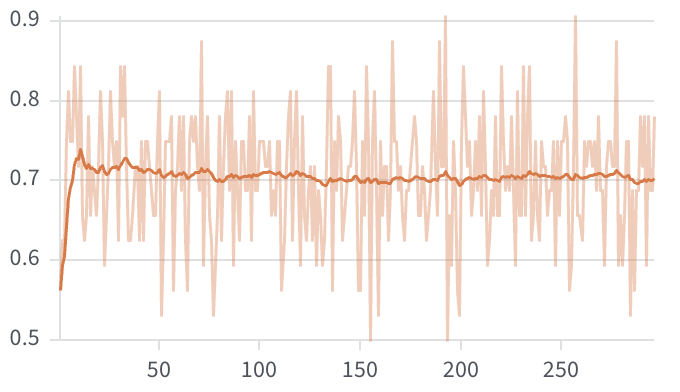} &
        \includegraphics[width=0.235\textwidth]{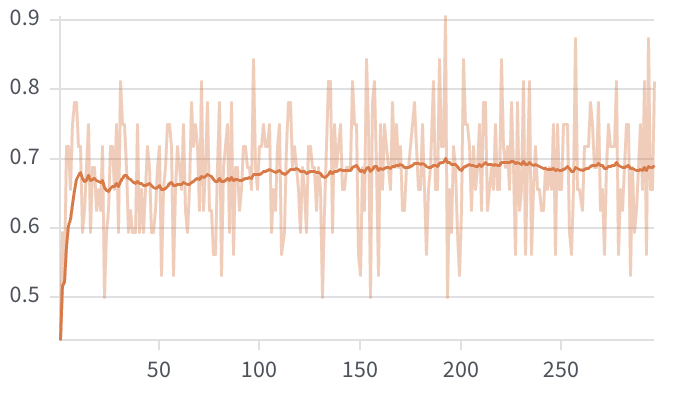}
        \\[-0.3em]

        $\mathrm{Acc}_{00}$ &
        $\mathrm{Acc}_{01}$ &
        $\mathrm{Acc}_{10}$ &
        $\mathrm{Acc}_{11}$

    \end{tabular}

    \caption{
    \textbf{Reliability-aware view weighting dynamics.}
    The top row shows the evolution of the learned harmonic-consensus
    weights for the four image--text views, and the bottom row reports
    the corresponding per-view pseudo-label accuracy. Each column
    corresponds to the same view $v_{it}$, enabling comparison between
    assigned consensus influence and empirical correctness during
    adaptation. The first index denotes the image state (original or
    transformed), and the second index denotes the text mode (direct
    solving or reframing).
    }
    \label{fig:reweighting_dynamics}
\end{figure*}

Figure~\ref{fig:reweighting_dynamics} shows the evolution of the four
view weights and their corresponding pseudo-label accuracies during
test-time adaptation on DynaMath with Qwen3-VL-8B-Instruct. The top row
shows the factorized harmonic-consensus weights
$w_{00},w_{01},w_{10},w_{11}$, while the bottom row reports the
post-hoc pseudo-label accuracy of each view. Ground-truth answers are
used only for evaluation and are unavailable during adaptation.

The view weights exhibit consistent alignment with the corresponding
pseudo-label accuracies. Views that maintain higher pseudo-label
accuracy generally retain larger consensus weights, while views with
lower accuracy receive reduced influence. This behavior suggests that
the pseudo-label-shift heuristic provides an effective label-free
estimate of view reliability, allowing the harmonic consensus mechanism
to adaptively balance original and transformed image--text views rather
than uniformly trusting all presentations.

The weighting dynamics remain smooth due to exponential reliability
mapping and temporal smoothing. Importantly, transformed views are not
permanently suppressed; they continue contributing when their
pseudo-label support remains stable. Although the heuristic cannot
guarantee that every update improves the instance-level ranking margin,
the theoretical analysis in Section~\ref{sec:weight_sensitivity}
characterizes when such reweighting increases harmonic ranking margins.
These empirical results demonstrate that the shared reliability factors
provide meaningful adaptive view weighting during test-time
optimization.

\end{document}